\documentclass[10pt,twocolumn,letterpaper]{article}

\usepackage{iccv}
\usepackage{times}
\usepackage{epsfig}
\usepackage{graphicx}
\usepackage{amsmath}
\usepackage{amssymb}
\usepackage{subcaption}
\usepackage{booktabs} 

\usepackage[breaklinks=true,bookmarks=false]{hyperref}

\iccvfinalcopy 

\def\iccvPaperID{****} 

\begin{document}

\title{ChatGPT Asks, BLIP-2 Answers:\\ Automatic Questioning Towards Enriched Visual Descriptions}

\author{Deyao Zhu \qquad Jun Chen$^*$ \qquad Kilichbek Haydarov$^*$ \\ 
Xiaoqian Shen \qquad Wenxuan Zhang \qquad Mohamed Elhoseiny\\
King Abdullah University of Science and Technology\\
\small\texttt{\{deyao.zhu, jun.chen, kilichbek.haydarov,}\\  
\small\texttt{xiaoqian.shen, wenxuan.zhang, mohamed.elhoseiny\}@kaust.edu.sa}\\
}

\maketitle

\begin{abstract}

Asking insightful questions is crucial for acquiring knowledge and expanding our understanding of the world. 
However, the importance of questioning has been largely overlooked in AI research, where models have been primarily developed to answer questions. 
With the recent advancements of large language models (LLMs) like ChatGPT, we discover their capability to ask high-quality questions when provided with a suitable prompt. 
This discovery presents a new opportunity to develop an automatic questioning system. 
In this paper, we introduce ChatCaptioner, a novel automatic-questioning method deployed in image captioning. 
Here, ChatGPT is prompted to ask a series of informative questions about images to BLIP-2, a strong vision question-answering model. 
By keeping acquiring new visual information from BLIP-2's answers, ChatCaptioner is able to generate more enriched image descriptions.
We conduct human-subject evaluations on common image caption datasets such as COCO, Conceptual Caption, and WikiArt, and compare ChatCaptioner with BLIP-2 as well as ground truth. 
Our results demonstrate that ChatCaptioner's captions are significantly more informative, receiving three times as many votes from human evaluators for providing the most image information.
Besides, ChatCaptioner identifies 53\% more objects within the image than BLIP-2 alone measured by WordNet synset matching.
Code is available at \url{https://github.com/Vision-CAIR/ChatCaptioner}
\end{abstract}


\let\thefootnote\relax\footnotetext{Preprint. $^*$Equal contribution.}

\section{Introduction}

\hspace{1mm}

\textit{``The important thing is not to stop questioning.''}

\qquad\qquad\qquad\qquad\qquad\qquad\qquad\qquad \textit{Albert Einstein}

\hspace{1mm}

\begin{figure}
  \centering
  \includegraphics[width=0.45\textwidth]{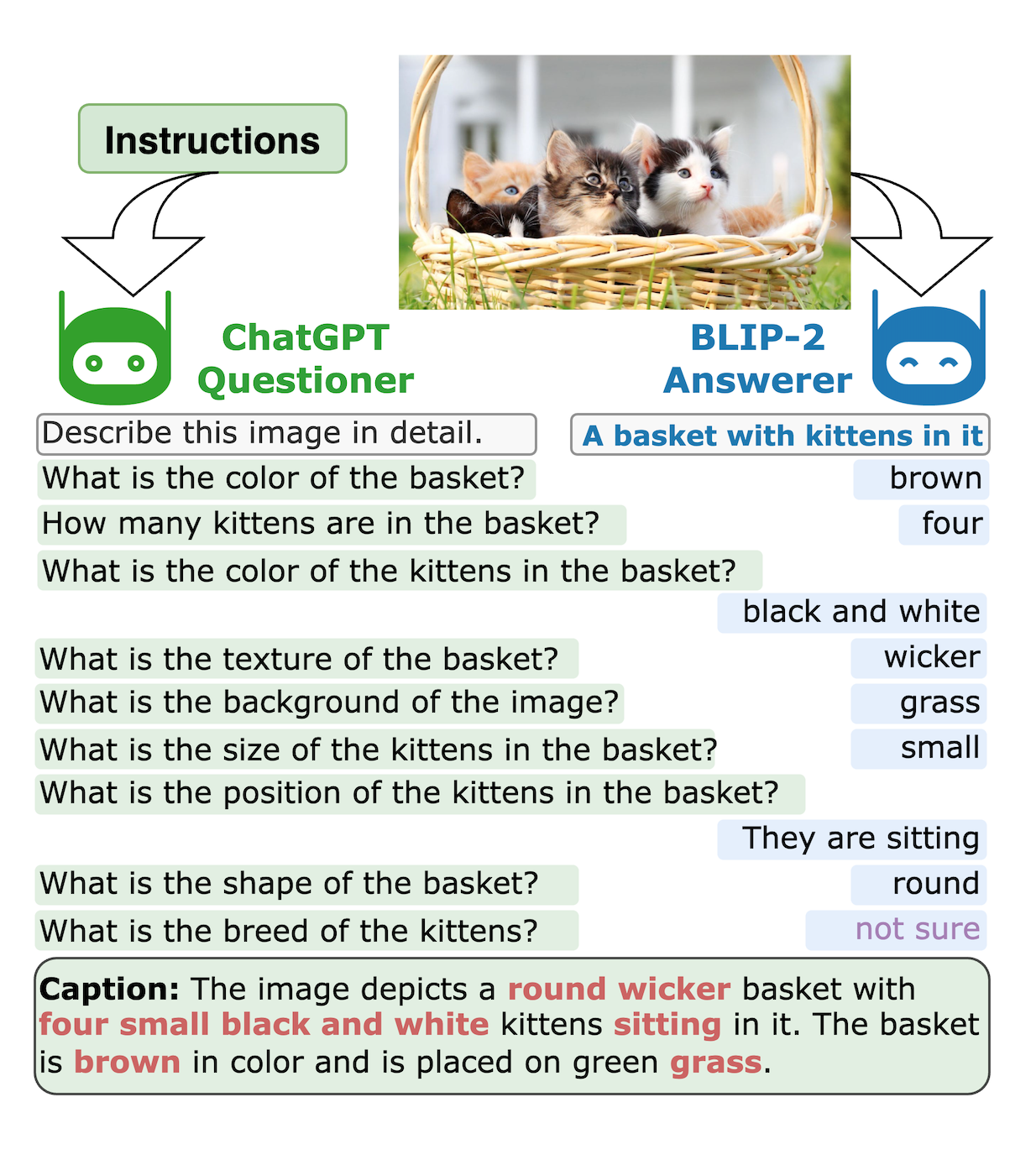}
  \caption{\textbf{Example of the dialog between ChatGPT and BLIP-2:} BLIP-2 fails to provide a detailed description in the first message exchange. More details about the image highlighted in red are obtained through multiple conversational interactions between the two models. Eventually, the questioner is able to produce a \textit{more detailed} caption about the image by focusing on multiple aspects of the image.}
  \label{fig:teaser}
  \vspace{-1em}
\end{figure}

Asking good questions is not only an essential component of effectively acquiring knowledge, but also plays a pivotal role in enhancing our intelligence and expanding our understanding of the world. 
Taking medical diagnoses as an example, doctors must ask patients targeted questions about their symptoms to gather relevant information and make accurate diagnoses. 
Likewise, in scientific research, asking insightful questions is paramount to advancing knowledge and discovering new findings that may have far-reaching implications.

However, the primary focus in recent AI research has been on developing models that can better answer questions, like InstructGPT \cite{ouyang2022training} in Open-Domain Question Answering \cite{yang2015wikiqa, rajpurkar2016squad, joshi2017triviaqa} and BLIP-2 \cite{li2023blip} in Visual Question Answering \cite{antol2015vqa, goyal2017making, hudson2019gqa}.
Despite the significant progress in the question-answering models, their effectiveness in providing useful information is heavily reliant on the quality of the questions they receive. 
In essence, these models depend on humans to ask insightful questions that can direct their generation of informative answers.
If we have an automatic questioning machine that keeps asking informative questions, the human questioners can be replaced and the question-answering models can be guided to provide more valuable knowledge automatically.

Recent studies \cite{wei2022finetuned, ouyang2022training, wei2022chain, kojima2022large} have highlighted the impressive zero-shot learning abilities of Large Language Models (LLMs) that are fine-tuned to follow instructions. 
These LLMs can perform new tasks in a zero-shot manner when presented with well-crafted instruction prompts.
We discover that such LLMs like ChatGPT \cite{openai2022chatgpt} have the ability to keep asking new and contextually relevant questions
when properly designed prompts are given. 
With this capability in place, building an effective automatic questioning machine is now a feasible task.

Based on our findings, we design an automatic questioning system on ChatGPT and integrate it into image captioning, where strong vision-language models like BLIP-2 \cite{li2023blip} are available to answer image-related questions.
Our method, named ChatCationer, generates more informative and enriched image captions by asking relevant questions to incrementally gain more information.
In detail, we design a prompting system that encourages ChatGPT to ask a series of informative questions that maximize its knowledge of the image, building on previous questions and answers.
Note that ChatGPT is a pure language model and cannot ``see'' any visual information.
We present the inquired image to BLIP-2 and set it as the question answerer.
At the end of the conversation, ChatGPT is prompted to summarize the discussion into a few sentences as the final enriched image description. 
An example of the conversation between ChatGPT and BLIP-2 and the final caption is shown in Fig.\ref{fig:teaser}.

We evaluate ChatCaptioner's captions on sampled images from COCO \cite{lin2014microsoft}, WikiArt \cite{saleh2015large}, and CC \cite{sharma2018conceptual} datasets based on the human subject evaluation experiment.
Compared to BLIP-2's state-of-the-art direct captioning performance, ChatCaptioner receives three times as many votes from human evaluators for providing richer image information. 
Besides, ChatCaptioner identifies 53\% more objects than BLIP-2 alone within the image.
Results verify the benefit of good questions to acquire more knowledge from existing AI models and the effectiveness of modern LLMs to serve as zero-shot automatic questioners.


\begin{figure*}
  \centering
  \includegraphics[width=0.9\textwidth]{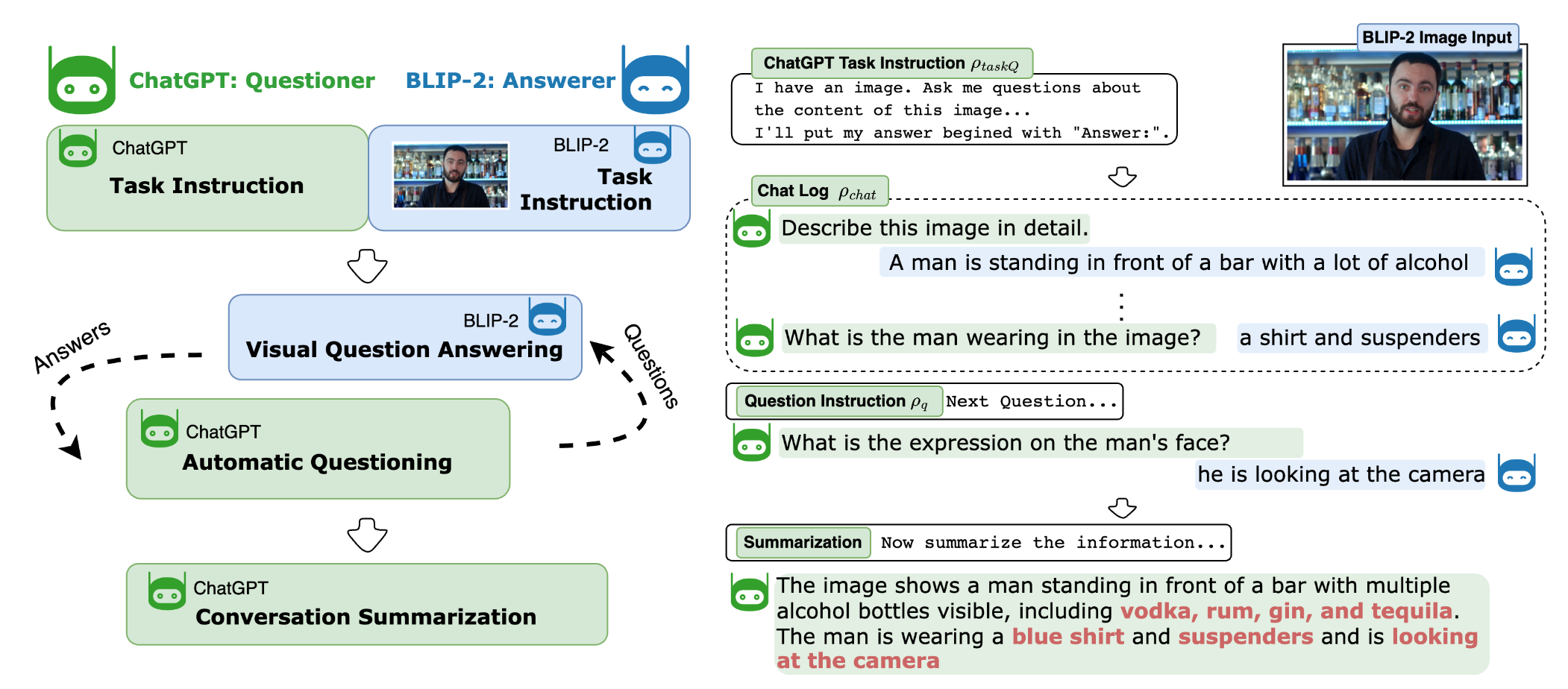}
  \caption{ChatCaptioner contains a questioner ChatGPT and an answerer BLIP-2. We start with a task instruction to set the context and outline the problem for ChatGPT. Once the answer from BLIP-2 is received, we prompt the chat log and question instruction to ChatGPT to continue asking information-acquiring questions. Finally, we provide a summarization instruction to ChatGPT to conclude the conversation as the image captions. Enriched details are highlighted in red.}
  \label{fig:overview}
\end{figure*}

\section{Related Works}
\paragraph{Learning to Ask Questions} 
Question generation \cite{mostow2009generating, heilman2010good} is the task of generating a question from a given passage and an answer.
Recent methods like \cite{jia2020ask, xiao2020ernie, liu2019learning, ghanem2022question, wang2022learning}
have explored different neural network architectures and training strategies for better performance.
However, in cases when we do not have the answer and need to ask questions for the answers, such methods are not applicable.
Visual Question Generation \cite{mostafazadeh2016generating, zhang2016automatic} is a task aimed at generating natural and engaging questions for a given image. 
Several works like \cite{patro-etal-2018-multimodal, Patro_2020_WACV, Li_2018_CVPR, jain2017creativity, vedd2021guiding, shen2019learning} have been proposed to solve this task. 
They focus on generating independent questions only and do not have the ability to keep asking new questions based on the previous questions.
Our work differs from previous studies significantly. 
First, we focus on acquiring more knowledge via the generated questions, instead of just generating them. 
Secondly, our method can keep asking new and relevant questions based on previously questions.
Third, our approach leverages modern large language models and requires zero training for questioning.

\paragraph{Large Language Model and Prompting}

Recent research \cite{brown2020language, kojima2022large, wei2022chain, wei2022emergent, wei2022finetuned, chung2022scaling, ouyang2022training} has revealed the abilities of Large Language Models (LLMs) like GPT-3 \cite{brown2020language} or PaLM \cite{chowdhery2022palm} to solve versatile tasks specified by prompting. For example, GPT-3 \cite{brown2020language} shows the capability to learn new tasks by providing a few task examples provided in the prompt, named in-context learning.
Moreover, Chain-of-Thought methods \cite{kojima2022large, wei2022chain} demonstrate that explicitly asking LLM to solve tasks step-by-step in the prompt improves the performance significantly.
Additionally, FLAN \cite{wei2022finetuned, chung2022scaling} demonstrates that LLMs with instruction tuning can accomplish new tasks in a zero-shot manner. 
Further studies, including InstructGPT \cite{ouyang2022training} and ChatGPT \cite{openai2022chatgpt}, show that the performance of LLMs can be enhanced even further by using reinforcement learning from human feedback \cite{christiano2017deep, stiennon2020learning}.
In our work, we leverage the instruction-following ability of LLMs and design prompts that enable ChatGPT to keep asking new questions about images.

\paragraph{Image Captioning and Visual Question Answering} 
Recent research in vision and language pertaining \cite{visualgpt, tsimpoukelli2021multimodal, alayrac2022flamingo, wang2022image, li2022blip, li2023blip} has advanced the performance for image captioning and visual question answering (VQA) by a large margin.
For example, VisualGPT \cite{visualgpt} shows the benefits of initialization with pretrained language models for more data-efficient training. 
Frozen \cite{tsimpoukelli2021multimodal} extend it by finetuning a vision encoder and aligning it with a frozen LLM. 
BEiT-3 \cite{wang2022image} and BLIP \cite{li2022blip} pretrain models using unified transformer architecture.
Flamingo \cite{alayrac2022flamingo} proposes a cross-attention design to align vision and language.
BLIP-2 \cite{li2023blip} introduces a lightweight Q-Former that converts visual features into tokens that can be directly understood by a frozen LLM, and achieves impressive results on both image captioning and VQA tasks.
In our work, our automatic questioning mechanism leverages the VQA capability of BLIP-2 to extract additional image information and enrich the image captions beyond the original BLIP-2 captions.

\section{ChatCaptioner}

In ChatCaptioner, we design an automatic questioning mechanism based on ChatGPT's zero-shot instruction-following ability to keep asking informative questions about images. 
BLIP-2, the vision-language model, then provides new image information according to the asked questions.
Finally, ChatGPT is prompted to summarize the chat history and generate the final image captions with rich details. An overview of our method is demonstrated in Fig.\ref{fig:overview}.

\subsection{Automatic Questioning}
To activate the questioning ability of ChatGPT, we design a prompting system that enables ChatGPT to generate questions based on previous chat logs.
Our prompting system for ChatGPT contains three components: 
a task instruction for explaining the task $\rho_{taskQ}$, 
a chat log to store previous questions and answers $\rho_{chat}$, 
a question instruction for generating high-quality questions $\rho_{q}$.
Each question is generated given the context $\rho_{taskQ} + \rho_{chat} + \rho_{q}$.
In addition, we design a question-trimming mechanism for the automatic post-processing of the generated questions.

\paragraph{ChatGPT Task Instruction $\rho_{taskQ}$}
The task instruction $\rho_{taskQ}$ sets the context and outlines the task that ChatGPT is required to perform. 
$\rho_{taskQ}$ directs ChatGPT to generate questions that aim to extract as much information as possible about an image. 
$\rho_{taskQ}$ is designed as follows:

\textit{I have an image. Ask me questions about the content of this image. Carefully asking me informative questions to maximize your information about this image content. Each time ask one question only without giving an answer. Avoid asking yes/no questions. I'll put my answer beginning with ``Answer:''.}

The prompt ``Each time ask one question only without giving an answer'' in $\rho_{taskQ}$ is designed to instruct ChatGPT to generate only one question per round. 
Without this prompt, ChatGPT may generate a set of questions at once, rather than generating them one by one in response to each answer.

\begin{figure*}
  \centering
    \includegraphics[width=0.95\textwidth]{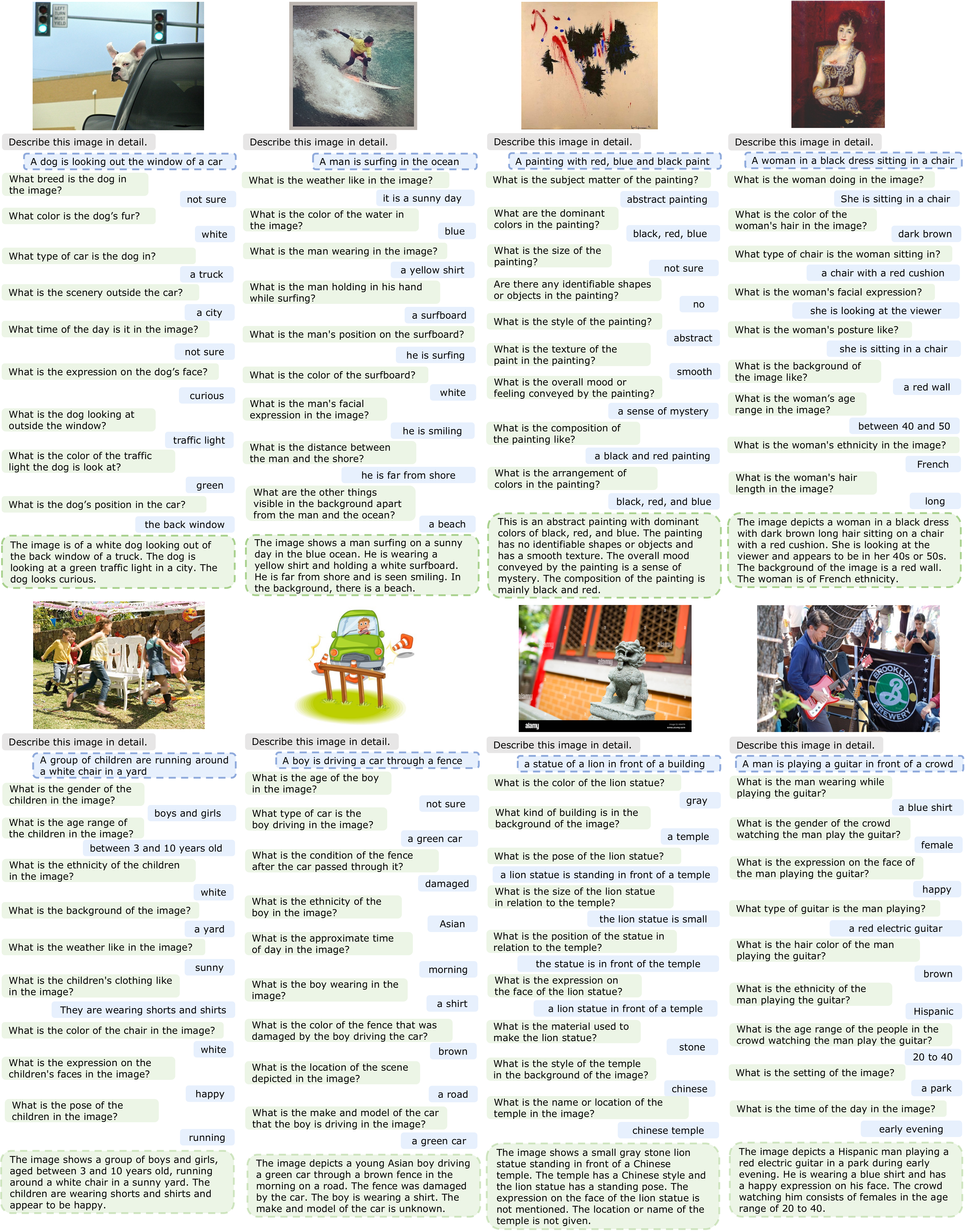}
  \caption{Qualitative examples of the chat log and the captions from ChatCaptioner in various images from COCO \cite{lin2014microsoft}, WikiArt \cite{saleh2015large}, and CC \cite{sharma2018conceptual}. Compared to the initial description from BLIP-2, questions from ChatGPT 
  extract additional image information and lead to enriched final captions.}
  \label{fig:Qualitative} 
\end{figure*}

\paragraph{Chat Log $\rho_{chat}$}
After the task instruction $\rho_{taskQ}$, we include a chat log $\rho_{chat}$ containing all the previous questions and answers. 
$\rho_{chat}$ follows a template for each Question-Answer round, which is as follows:

\textit{Question: $\langle question \rangle$ Answer: $\langle answer \rangle$}

Note that we hard-code the first question as ``Describe the image in detail'' to start the conversation.
Based on the first response of BLIP-2, which provides a brief initial description of the image, ChatGPT is prompted to ask follow-up questions to extract more information about the image.

\paragraph{Question Instruction $\rho_{q}$}
To guide ChatGPT in generating new questions, we provide a question instruction $\rho_{q}$ before each question generation. 
$\rho_{q}$, located after the chat log, cues ChatGPT to generate a new question and aims to ensure that the questions are of high quality. It's designed as follows:

\textit{Next Question. Avoid asking yes/no questions. Question:}

The prompt ``Next Question'' in $\rho_{q}$ is critical to ensure that ChatGPT continues to ask questions. 
Without it, ChatGPT may produce undesired sentences after a few Question-Answer rounds. 
Additionally, we notice that ChatGPT prefers asking yes/no questions which are usually not as informative as other questions. We therefore add the prompt ``Avoid asking yes/no questions'' to reduce the generation of yes/no questions.

\paragraph{Question Trimming}
Despite our explicit instruction to not answer the question itself, we observe that sometimes ChatGPT fabricates an answer after asking the question. Fortunately, we find that these fabricated answers always begin with the text ``Answer:'', following the template specified in the prompt. Therefore, we automatically remove these fabricated answers by discarding the generated text starting from ``Answer:''.

\subsection{Question Answering}
Similar to ChatGPT, our BLIP-2 prompting mechanism consists of three components: a task instruction $\rho_{taskA}$, the chat log $\rho_{chat}$ same as the ChatGPT one, and an answer instruction $\rho_{a}$.
Each answer generation is prompted by $\rho_{taskA} + \rho_{chat} + \rho_{a}$.
Also, we have an answer-trimming mechanism for post-processing.

\paragraph{BLIP-2 Task Instruction $\rho_{taskA}$}
We design the BLIP-2 task instruction $\rho_{taskA}$ to alleviate the issue of hallucinating non-existent information in the image. 
$\rho_{taskA}$ includes an uncertainty prompt ``If you are not sure about the answer, say you don't know honestly'' that encourages BLIP-2's honest admission of lack of knowledge.
The instruction is as follows:

\textit{Answer given questions. If you are not sure about the answer, say you don't know honestly. Don't imagine any contents that are not in the image.}

\paragraph{Answer Instruction $\rho_{a}$}
After the chat log $\rho_{chat}$, we provide a straightforward answer instruction to guide BLIP-2's answering process. The instruction is structured as follows:

\textit{Question: $\langle question \rangle$ Answer:}

\paragraph{Answer Trimming} 
Similar to ChatGPT, BLIP-2 occasionally generates a question after providing an answer.
As the LLM backend of BLIP-2, the FLAN-T5 model \cite{chung2022scaling}, has a much weaker questioning ability than ChatGPT shown later in Sec.\ref{sec: question analysis}, we automatically filter out these questions by discarding any texts starting with ``Question:''.

\subsection{Context Summarizing}
To obtain a concise summary of the conversation between ChatGPT and BLIP-2 as the final image caption, we use a summarization instruction after the conversation. 
This instruction, located after the chat log, prompts ChatGPT to generate a summary using the following structure:

\textit{Now summarize the information you get in a few sentences. Ignore the questions with answers no or not sure. 
Don't add information. Don't miss information. Summary:}

\begin{figure}[t]
  \centering
    \includegraphics[width=0.45\textwidth]{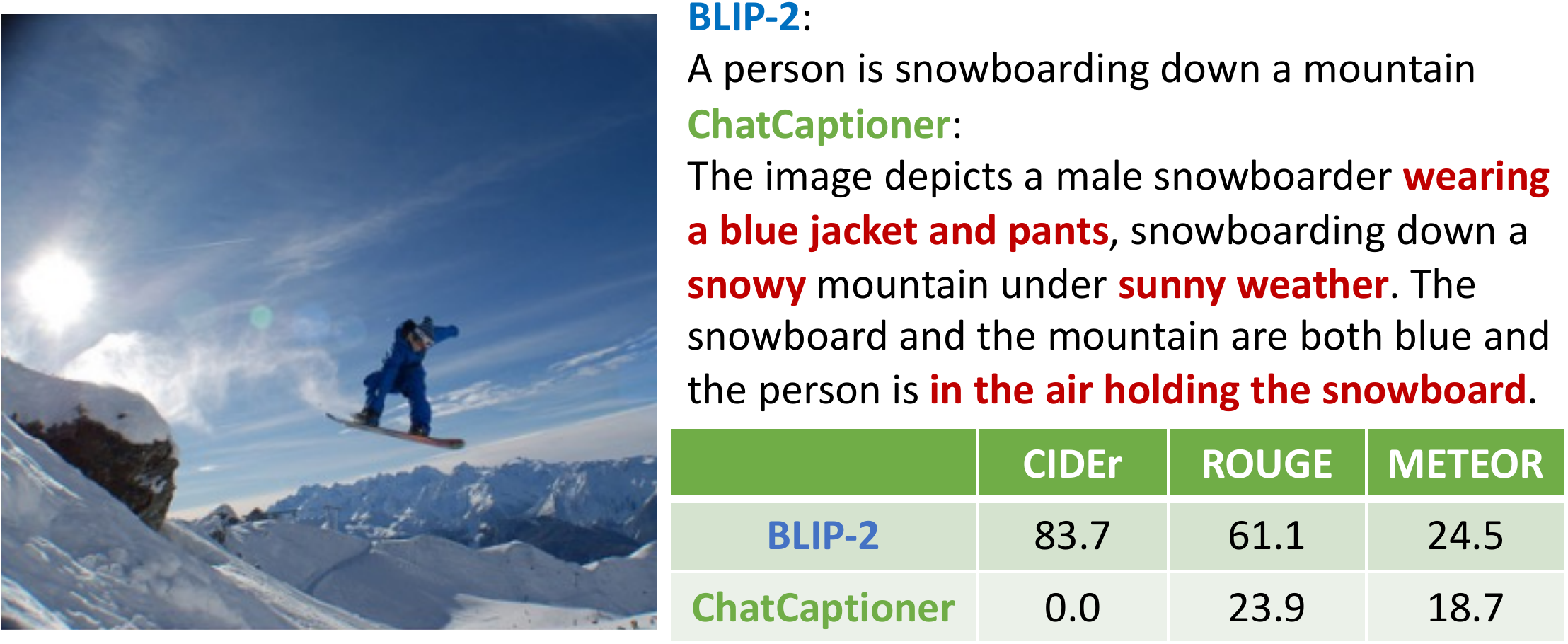}
  \caption{An example of the limitation of traditional image caption metrics. Although ChatCaptioner extracts more image details highlighted in red compared to BLIP-2, it receives much lower scores under traditional metrics.
  }
  \label{fig:metric_issue}
\end{figure}

\section{Experiments}

In this section, we explore ChatCaptioner from various perspectives through a series of experiments, including an assessment of the informativeness and accuracy of its captions, as well as an analysis of the quality of its generated questions. 
Qualitative results including chat logs and final captions on various images from different datasets are shown in Fig.\ref{fig:Qualitative}.

\noindent \textbf{Details of Model Deployment.}
For our experiments, we use the ChatGPT model \textit{``gpt-3.5-turbo''} available on the OpenAI API \cite{chatgptapi}. This model is the most powerful GPT-3.5 model accessible through the API during our project. 
For BLIP-2, we use the biggest version containing a FLAN-T5 \cite{chung2022scaling} language model with 11 billion parameters and a ViT-G/14 model from EVA-CLIP \cite{fang2022eva}.
In all experiments, BLIP-2 answers 10 questions per image, with the first question being hard-coded as \textit{``Describe the image in detail.''}. The remaining 9 questions are from ChatGPT, unless otherwise specified.

\noindent \textbf{Limitation of Traditional Metrics.}
The conventional image captioning metrics such as\cite{vedantam2015cider}, ROUGE \cite{lin2002manual}, and METEOR \cite{banerjee2005meteor} are frequently employed to measure the quality of captions. 
However, the usage of these metrics to evaluate ChatCaptioner can be limiting, because these metrics are designed to measure the similarity between the tested caption and reference captions, assuming that the reference captions are perfect image descriptions. 
Since ChatCaptioner generates captions that contain a greater level of detail than the reference captions, the metrics may yield low similarity scores, leading to inferior overall performance scores compared to other image captioning approaches like BLIP-2. 
This issue is depicted in Fig. \ref{fig:metric_issue}. Thus, in our experiments, we primarily rely on human assessments to comprehensively analyze the performance of ChatCaptioner from various perspectives.
The design of all the human evaluation interfaces is presented in Appendix.

\subsection{Information Analysis}

\paragraph{Does ChatCaptioner extract more information from the image?}
We design an experiment to evaluate whether ChatCaptioner is able to generate captions with more information about the image than BLIP-2 alone and ground truth captions. 
We randomly selected 100 photos from the COCO \cite{lin2014microsoft} validation set, 100 artworks from WikiArt \cite{saleh2015large} dataset with ground truth captions from ArtEmis \cite{achlioptas2021artemis}, and 100 internet images from the CC \cite{sharma2018conceptual} validation dataset. 
Human evaluators on Amazon Mechanical Turk are presented with an image and four captions - one from our method, one from BLIP-2, one ground truth caption, and one fake caption for quality control. 
Evaluators are asked to pick the caption that offers the richest information about the image.
Results are demonstrated in Tab.\ref{tab:info}.
On average, ChatCaptioner receives three to four times as many votes as pure BLIP-2's captions and ground truth captions, demonstrating the effectiveness of using an Automatic Question-Asking machine to enhance image information extraction from a VQA model.

\begin{table}[t]
\centering
\caption{Human votes on the captions containing the most image information.}
\label{tab:info}
\scalebox{0.7}{
\begin{tabular}{ccccc}
\toprule
\textbf{Methods} & \textbf{COCO} & \textbf{WikiArt} & \textbf{CC}  & \textbf{Avg.} \\
\midrule
GT & 26\% & 14\% & 8.5\%  & 16.2\% \\
BLIP-2 & 21\% & 12.5\% & 23\% & 18.8\% \\
\textbf{Ours} & \textbf{53\%} & \textbf{73.5\%} & \textbf{68.5\%} & \textbf{65\%} \\
\bottomrule
\end{tabular}
}
\vspace{-1em}
\end{table}

\noindent \textbf{How many objects in images can ChatCaptioner discover?}
Here, we randomly sampled 200 images from the Pascal VOC \cite{Everingham10} dataset and considered all class labels included in the segmentation masks as the ground truth objects. We then assessed how many of these objects are included in the captions.
We utilize WordNet from the Natural Language Toolkit (NLTK) \cite{bird2009natural} to find words with similar semantic meanings. Specifically, we match two words if the Wu-Palmer Similarity of their synsets is greater than 0.9 or if one word's synset is included in the other's closure.
Tab.\ref{tab:obj_cover} presents the experimental results, where 1154 objects are identified in the 200 sampled images. 
BLIP-2 covers only 383 of them, while with the help of automatic questioning, ChatCaptioner increases the coverage by 53\% to 586, suggesting that the automatically generated questions help BLIP-2 find more objects in the images.

\begin{table}[t]
\centering
\caption{Numbers of objects discovered by captions.}
\label{tab:obj_cover}
\scalebox{0.7}{
\begin{tabular}{cccc}
\toprule
\textbf{Methods} & \textbf{Covered/All} &  \textbf{Ratio} & \textbf{Improved}\\
\midrule
BLIP-2 & 383/1154 & 33.2\% & - \\
\textbf{Ours} & \textbf{586}/1154 & \textbf{50.8\%} & \textbf{53.0\%}\\
\bottomrule
\end{tabular}
}
\vspace{-0.5em}
\end{table}

\subsection{Correctness Analysis}

\paragraph{How accurate are the captions from ChatCaptioner?}

To evaluate the correctness of ChatCaptioner's captions, we conducted a human evaluation where evaluators were presented with an image and a generated caption, as well as all questions and answers between ChatGPT and BLIP-2. The evaluators were asked to verify the correctness of the caption with respect to the image, select any incorrect answers from BLIP-2, and judge whether the incorrectness of the caption can be attributed to the wrong answers. The experiments were performed on samples from COCO \cite{lin2014microsoft}, WikiArt \cite{saleh2015large}, and CC \cite{sharma2018conceptual} datasets, similar to previous experiments, and each image was evaluated by 4 different evaluators. Results are presented in Tab.\ref{tab:correct}.
Our findings reveal that approximately 80\% of the generated captions are deemed correct. Moreover, BLIP-2 is able to provide correct answers to around 67\% of the questions asked by ChatGPT. Among the incorrect captions, 94\% of them are caused by BLIP-2's wrong answers, suggesting that BLIP-2 is the primary source of incorrectness. This implies that using a more powerful visual question-answering (VQA) model may help to enhance the overall performance of the system in the future.

\begin{table}[t]
\centering
\caption{Correctness Analysis of ChatCaptioner. Overall, BLIP-2 can correctly answer about 66.7\% of ChatGPT's questions. 81\% of the final captions are deemed correct by human evaluators. Besides, 94\% of the wrong captions are caused by BLIP-2's wrong answers.}
\label{tab:correct}
\scalebox{0.7}{
\begin{tabular}{ccccc}
\toprule
\textbf{} & \textbf{COCO} & \textbf{WikiArt} & \textbf{CC}  & \textbf{Avg.} \\
\midrule
Answer Correct Rate & 64\% & 73\% & 63\%  & 66.7\%\\
Caption Correct Rate & 77\% & 78\% & 88\% & 81\% \\
Issues From BLIP-2 & 100\% & 82\% & 100\% & 94\% \\
\bottomrule
\end{tabular}
}
\vspace{-1em}
\end{table}

\begin{figure}
  \centering
    \includegraphics[width=0.45\textwidth]{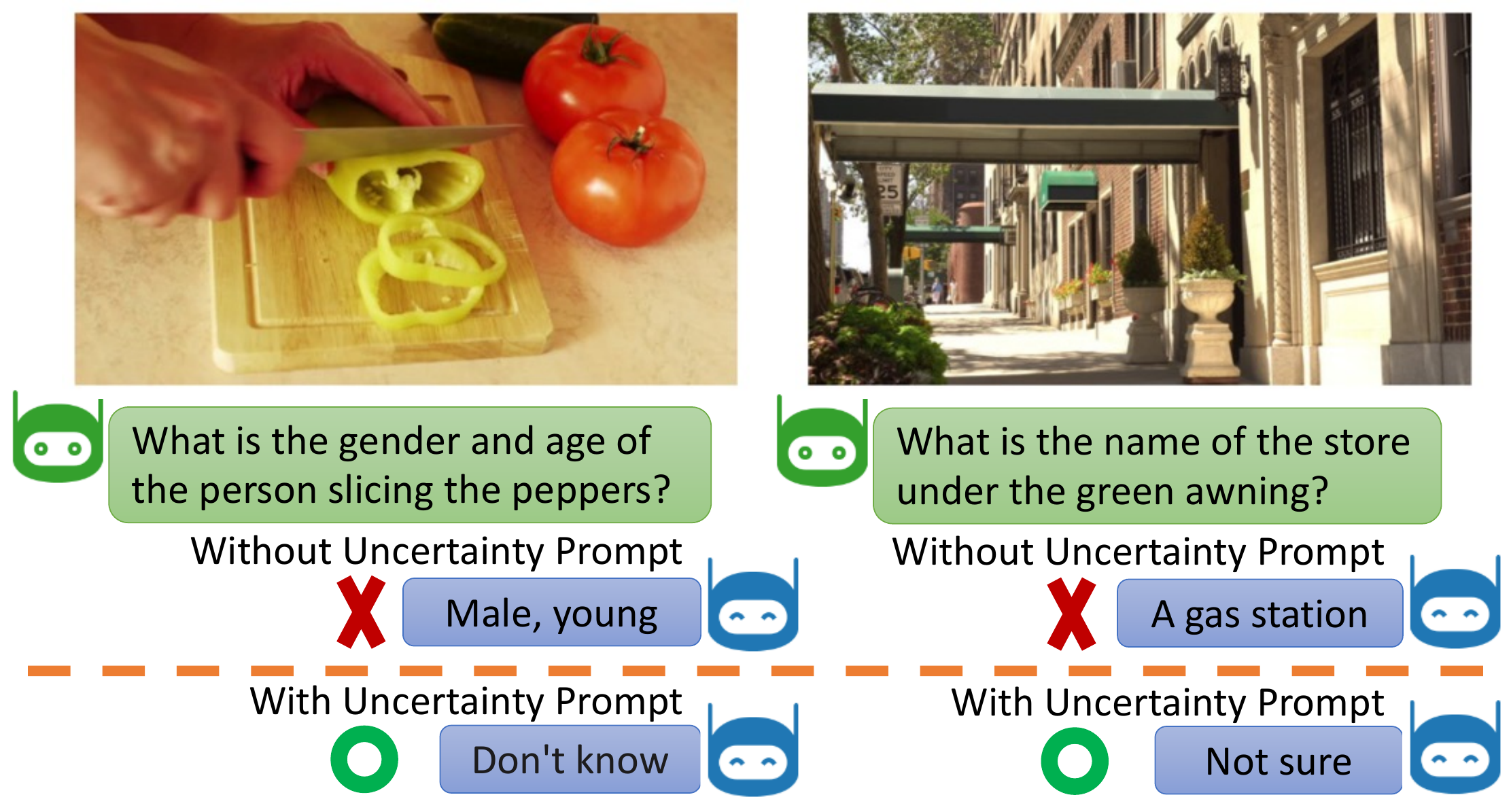}
  \caption{Examples of BLIP-2's answers with and without the uncertainty prompt. 
  The uncertainty prompt helps BLIP-2 avoid making up an answer when it encounters questions that cannot be answered based on the image.
  }
  \label{fig:uncertainty}
\end{figure}

\noindent\textbf{Does BLIP-2 know it doesn't know?}
BLIP-2 usually makes up answers if the question cannot be answered based on the given image. In other words, BLIP-2 doesn't know that it doesn't know this information.
To mitigate this issue, we incorporate an uncertainty prompt \textit{``If you are not sure about the answer, say you don't know honestly.''} in our BLIP-2 task instruction $\rho_{taskA}$. This prompt encourages the model to say it doesn't know when it is not confident in its response.
Two examples are demonstrated in Fig.\ref{fig:uncertainty}.
In the first example, BLIP-2 is presented with an image that only shows two hands and is asked to determine the gender and age of the person.
With the uncertainty prompt, BLIP-2 changes its answer from guessing a young male to honestly saying ``Don't know''.
In the second example, BLIP-2 initially thought that the unrecognizable store in the photo was a gas station, but with the addition of the uncertainty prompt, it changes the answer to ``Not sure''.
More examples can be found in Fig.\ref{fig:Qualitative} and Appendix.

\begin{table}[t]
\centering
\caption{Analysis on questions that BLIP-2 is unsure about. 60\% deemed unanswerable by humans. 20\% cannot be correctly answered by BLIP-2. Overall, BLIP-2 makes mistakes on 80\% of these questions.}
\label{tab:uncertainty}
\scalebox{0.75}{
\begin{tabular}{ccccc}
\toprule
\textbf{} & Total Uncertain & Unanswerable & Answerable & Avoided Bad \\
\textbf{} & Questions & Questions & But Wrong & Answers \\
\midrule
Num. & 147 & 88 & 30 & 118 \\
Ratio & - & 60\% & 20\% & 80\% \\
\bottomrule
\end{tabular}
}
\vspace{-1em}
\end{table}

\vspace{1.5em}
\noindent \textbf{How effective is the uncertainty prompt?}
To investigate whether the questions that BLIP-2 is unsure about can be answered by humans, we randomly selected 200 images from the CC \cite{sharma2018conceptual} dataset and collected 1,800 questions based on these images. We then identify 147 questions that BLIP-2 is uncertain about, present these questions to human evaluators, and ask them to answer based on the image content. 
Results presented in Tab.\ref{tab:uncertainty} demonstrate that approximately 60\% of these questions are deemed unanswerable based on the image content. 
For the remaining answerable questions, BLIP-2 cannot correctly answer 30 of them.
In total, without the uncertainty prompt, BLIP-2 will generate 118 incorrect answers out of 147 uncertain questions, resulting in an error rate of approximately 80\%.
In addition, out of the original 1800 questions, BLIP-2 has 674 wrong answers. 
Taking the 147 potential wrong answers avoided by the uncertainty prompt into account, 
the uncertainty prompt reduces about 15\% of the wrong answers.

\subsection{Question Analysis}
\label{sec: question analysis}

\begin{figure}
  \centering
    \includegraphics[width=0.47\textwidth]{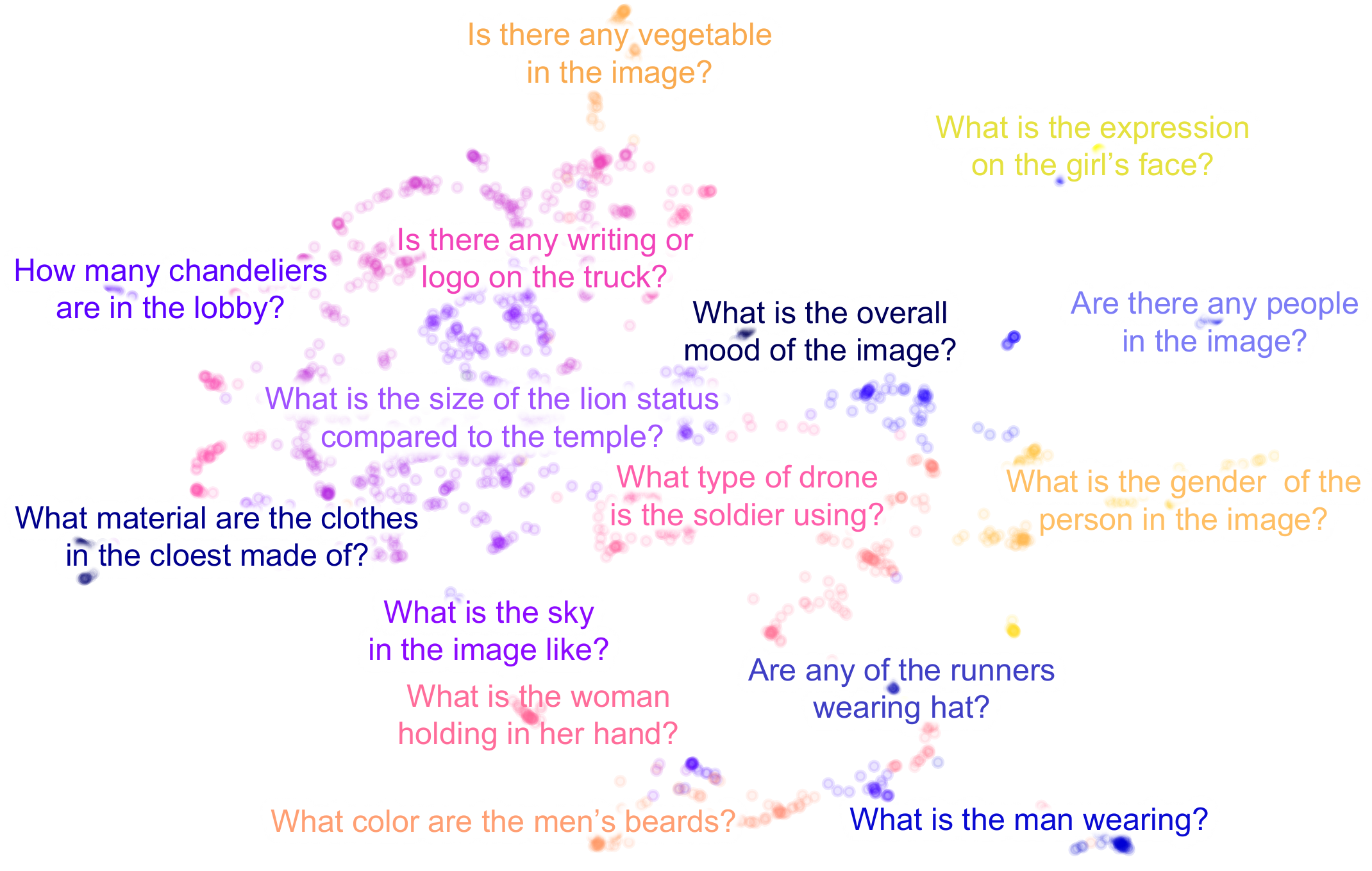}
  \caption{Visualization of question embeddings and question examples printed at the corresponding positions. Our method can ask diverse questions focusing on various perspectives of the image.}
  \label{fig:q_div}
    \vspace{-1em}
\end{figure}

\noindent \textbf{How diverse are the automatic generated questions?}
To investigate whether ChatGPT-based Automatic Question Asking mechanism can produce diverse questions, we analyze 1800 generated questions on 200 random image samples from the CC \cite{sharma2018conceptual} dataset.
After filtering out the repeated ones, we have in total of 1419 unique questions.

To visualize the diversity of the generated questions, we compute the sentence embedding of each question using Sentence-BERT \cite{reimers2019sentence} and reduce their dimensions to two for visualization using UMAP \cite{mcinnes2018umap}. The resulting visualization is presented in Fig.\ref{fig:q_div} with question examples plotted at corresponding positions.

Our analysis shows that the questions generated by ChatCaptioner are diverse and cover various perspectives, including high-level information like \textit{``What is the overall mood of this image?''},  object properties like \textit{``What material are the clothes in the closet made of?''}, numbers of objects like \textit{``How many chandeliers are in the lobby?''}, facial expression like \textit{``What is the expression on the girl's face?''}, environmental information like \textit{``What is the sky in the image like?''}, and so on. 

Overall, our findings suggest that our ChatGPT-based automatic questioning mechanism is able to generate diverse questions, explore various aspects of the image, and acquire more informative responses.

\begin{figure}
  \centering
    \includegraphics[width=0.47\textwidth]{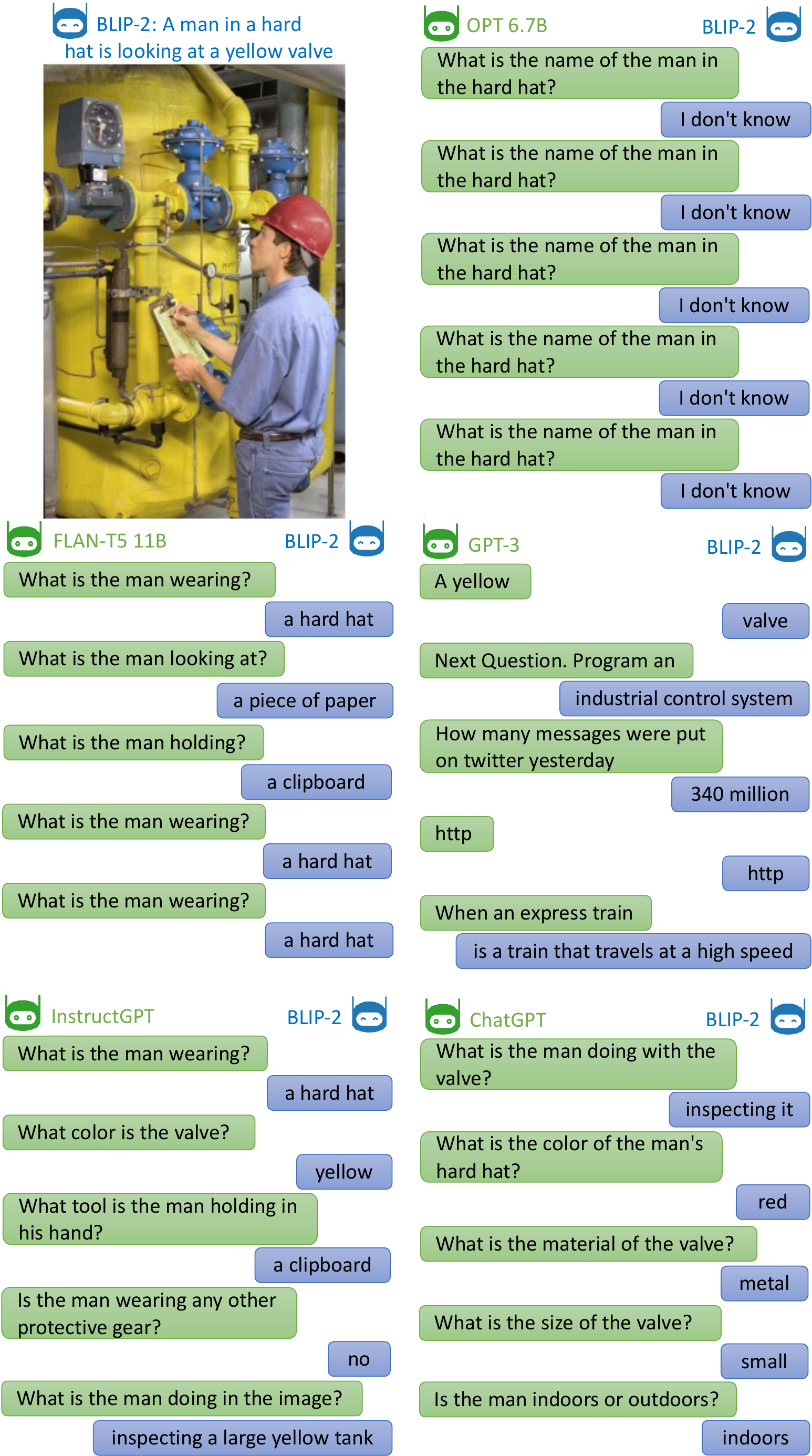}
  \caption{Examples of questions from various language models based on initial image descriptions. ChatGPT and InstructGPT demonstrate their ability to ask new and informative questions automatically. In contrast, FLAN-T5 11B and OPT 6.7B tend to repeat old questions, indicating weaker questioning abilities. The original GPT-3 struggles the most and fails to ask any related questions.}
  \label{fig:diff_model}
\end{figure}

\noindent \textbf{Can we use other LLMs as questioners?}
In addition to ChatGPT, we evaluate the automatic questioning capabilities of several other large language models (LLMs), including InstructGPT \cite{ouyang2022training}, the original GPT-3 \cite{brown2020language} without instruction finetuning, FLAN-T5 11B \cite{chung2022scaling} (11B parameters), and OPT 6.7B \cite{zhang2022opt} (6.7B parameters).
Fig.\ref{fig:diff_model} showcases the different question-asking abilities given the initial description from BLIP-2. 
We observe that InstructGPT shows a similar question-asking ability as ChatGPT and is able to generate meaningful and diverse questions.
While FLAN-T5 11B and OPT 6.7B also exhibits the ability to ask questions, it repeats a single question several times.
Interestingly, the original GPT-3 model shows the worst question generation ability and fails to generate any meaningful questions related to the image. 

We further analyze their 1800 generated questions on 200 CC \cite{sharma2018conceptual} samples, at a rate of 9 questions per image.
We skip the original GPT-3 model here as it is not able to generate any meaningful questions in our early exploration.
The tested LLMs' questioning abilities are evaluated in two ways. 
Firstly, we measure the average number of unique questions per image, which indicates whether the model can keep asking new questions in a single dialogue. 
Secondly, we count the unique questions over all the 1800 asked questions to see if the questioner could customize the questions according to the given contexts or just ask fixed predefined questions.

Our findings, as shown in Tab.\ref{tab:diff_model}, reveal that ChatGPT and InstructGPT almost never repeat their question in a single nine-question dialogue and generate a total of around 1400 unique questions out of 1800, suggesting that they are able to ask new questions according to the contexts.
In contrast, FLAN-T5 11B and OPT 6.7B have a tendency to repeat old questions, averaging about only 2 unique questions per image and generating less than 170 unique questions in total.
Our study suggests that to develop a questioning machine that can automatically generate novel and customized questions, it may be necessary to utilize LLMs with at least dozens of billions of parameters that are specifically fine-tuned for improved instruction following ability.

\begin{table}[t]
\centering
\caption{Number of unique questions per dialogue and in total. InstructGPT and ChatGPT excel at generating diverse questions, rarely repeating questions within a dialogue, and outperforming OPT 6.7B and FLAN-T5 11B.}
\label{tab:diff_model}
\scalebox{0.75}{
\begin{tabular}{ccccc}
\toprule
Unique Q/Total Q & OPT 6.7B & FLAN-T5 & InstructGPT & ChatGPT \\
\midrule
Per Dialogue & 1.75/9 & 2.03/9 & 9/9 & 8.98/9 \\
All Questions & 166/1800 & 169/1800 & 1400/1800 & 1419/1800 \\
\bottomrule
\end{tabular}
}
\end{table}

\subsection{Limitation}

The caption correctness of ChatCaptioner relies on the answers from BLIP-2. 
Although we design the uncertainty prompt to reduce the number of wrong answers from BLIP-2, a small portion of the answers are still incorrect.
Combining automatic questioning with better vision-language models in the future may enhance its visual  description ability.
As ChatCaptioner is based on LLMs, it also inherits the risks of LLM and might sometime generate offensive or socially biased conversations and captions.
Finetuning the system with a filtered dataset or human feedback may alleviate this issue.

\section{Conclusion}

In this work, we discover that advanced large language models possess the ability to pose insightful and diverse questions when provided with well-crafted prompts.
Based on our findings, we develop an automatic questioning system named ChatCaptioner for the task of image captioning. 
By prompting ChatGPT to keep asking questions that expand its understanding of an image, ChatCaptioner guides BLIP-2 to provide comprehensive image information, resulting in image captions that are significantly more detailed and enriched.
ChatCaptioner demonstrates the power of automatic questioning systems to effectively extract desired information. 
Through our work, we aim to draw attention to the potential of automatic questioning systems in AI and inspire further research in various domains.

{\small
\bibliographystyle{ieee_fullname}
\bibliography{egbib}
}

\newpage
\newpage
\onecolumn
\section{Appendix}

\subsection{Cost}
Our method is based on the ChatGPT model, specifically the \textit{gpt-3.5-turbo} version which we access through OpenAI's API. 
At the time of our project, the cost for using 1000 tokens in \textit{gpt-3.5-turbo} was 0.002 US Dollars. 
On average, we spent approximately 2500 tokens for each image for ten Question-Answer rounds, which translates to a cost of approximately 0.005 US Dollars per image.

\subsection{Yes/No Question Ablation}
Usually, yes/no questions contain relatively less information.
To reduce the generation of yes/no questions from ChatGPT, we explicitly add a prompt \textit{``Avoid asking yes/no questions''} in the task instruction $\rho_{taskQ}$ and the question instruction $\rho_{q}$. Our ablation study in Tab.\ref{tab:yes_no} shows that this prompt reduces the generation of yes/no questions from 33\% of the cases to 2\% in 1800 questions on 200 random CC \cite{sharma2018conceptual} samples, verifying its effectiveness.

\begin{table}[h]
\centering
\caption{Effectiveness of the yes/no prompt.}
\label{tab:yes_no}
\scalebox{0.75}{
\begin{tabular}{ccccc}
\toprule
\textbf{} & Total Question & Yes/No Question w/o Prompt & Yes/No Question with Prompt \\
\midrule
Num. & 1800 & 595 & 38 \\
Ratio & - & 33\% & 2\% \\
\bottomrule
\end{tabular}
}
\vspace{-1em}
\end{table}

\newpage
\subsection{Human Evaluation Interface}

\begin{figure}[b]
\centering
\begin{subfigure}{0.45\textwidth}
  \includegraphics[width=\textwidth]{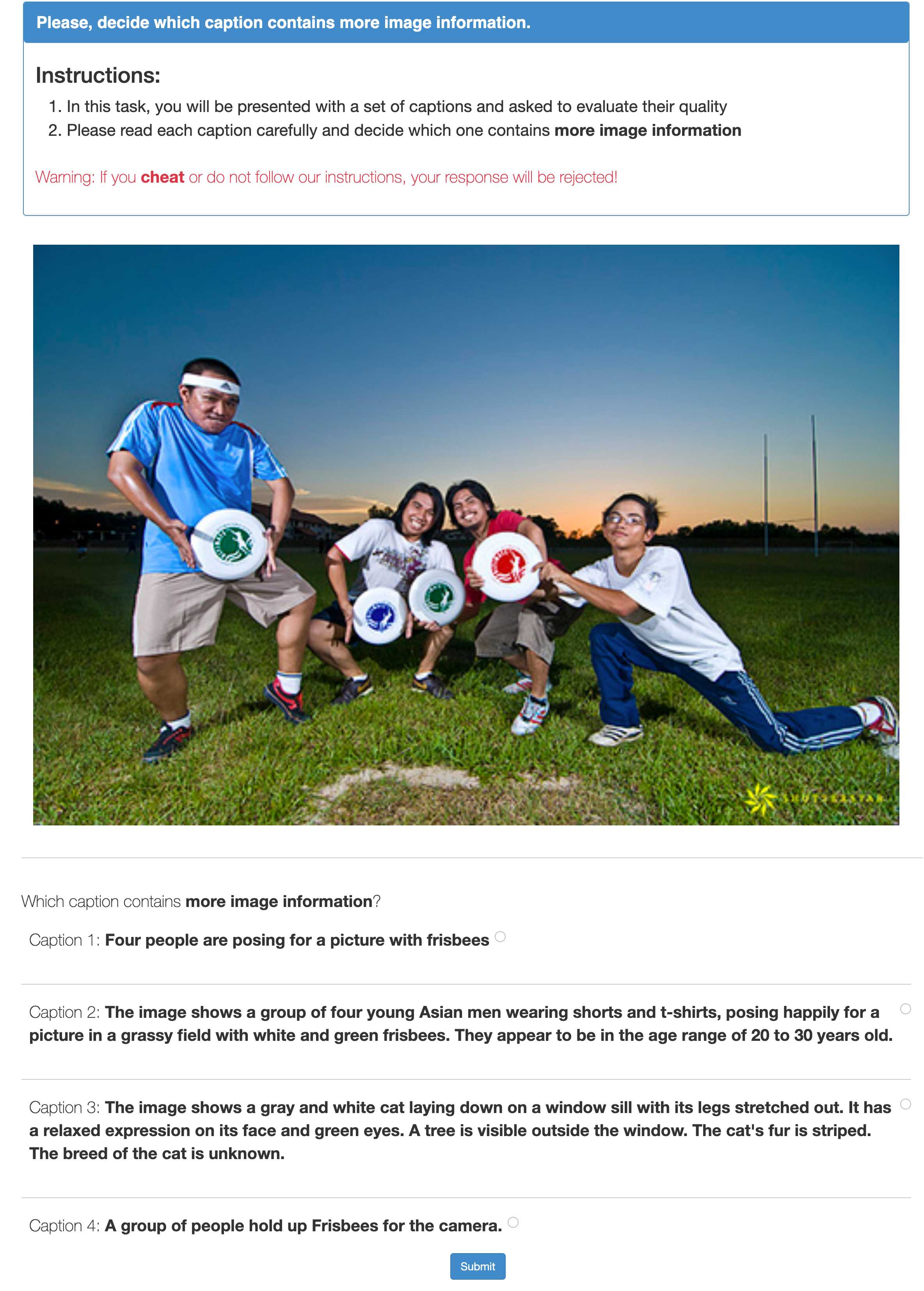}
  \caption{Human evaluation interface of the information experiments in Tab.\ref{tab:info}.}
\end{subfigure}
\hspace{5mm}
\begin{subfigure}{0.4\textwidth}
  \centering
  \includegraphics[width=\textwidth]{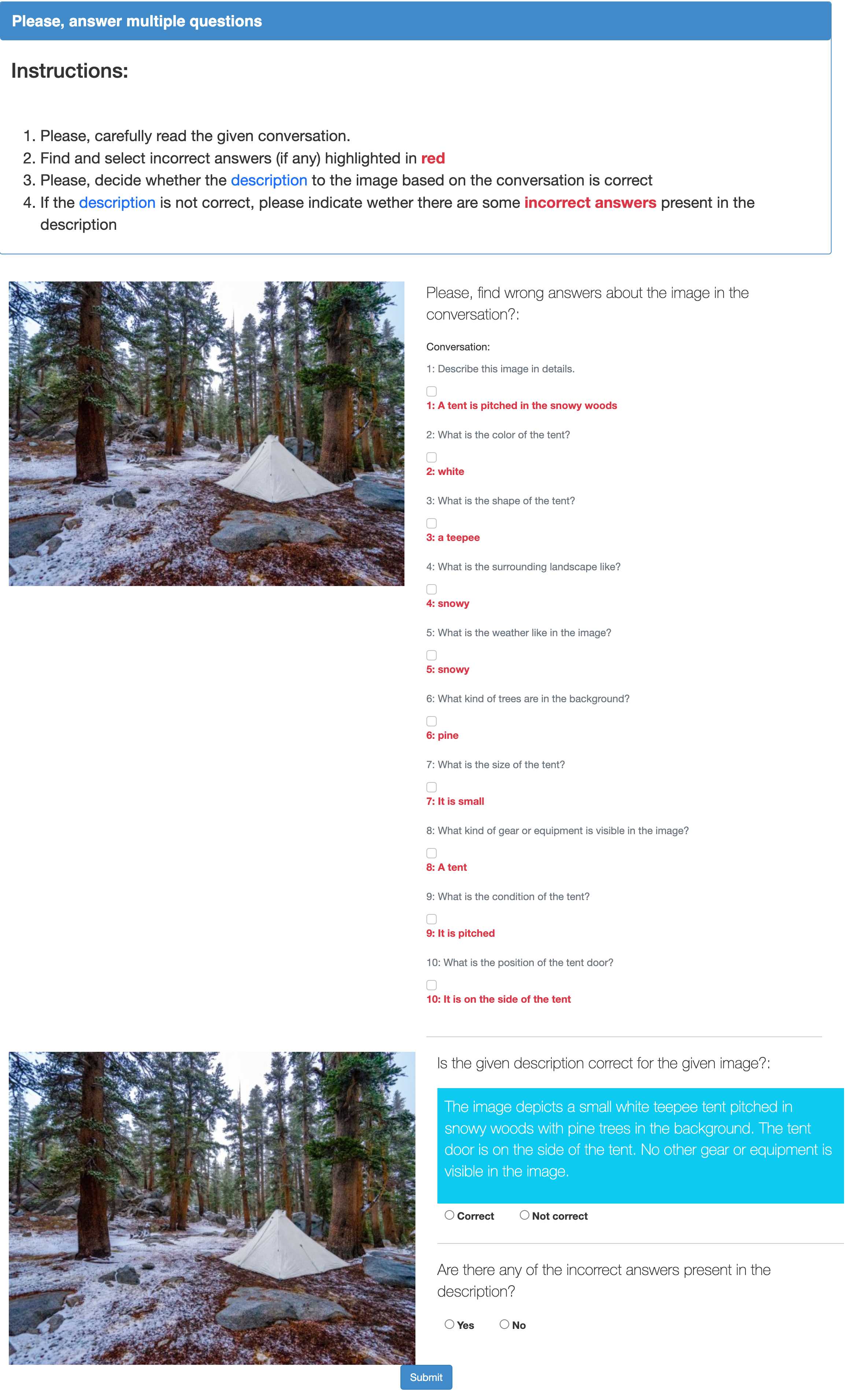}
  \caption{Human evaluation interface of the correctness experiments in Tab.\ref{tab:correct}.}
\end{subfigure}
\begin{subfigure}{0.4\textwidth}
  \centering
  \includegraphics[width=\textwidth]{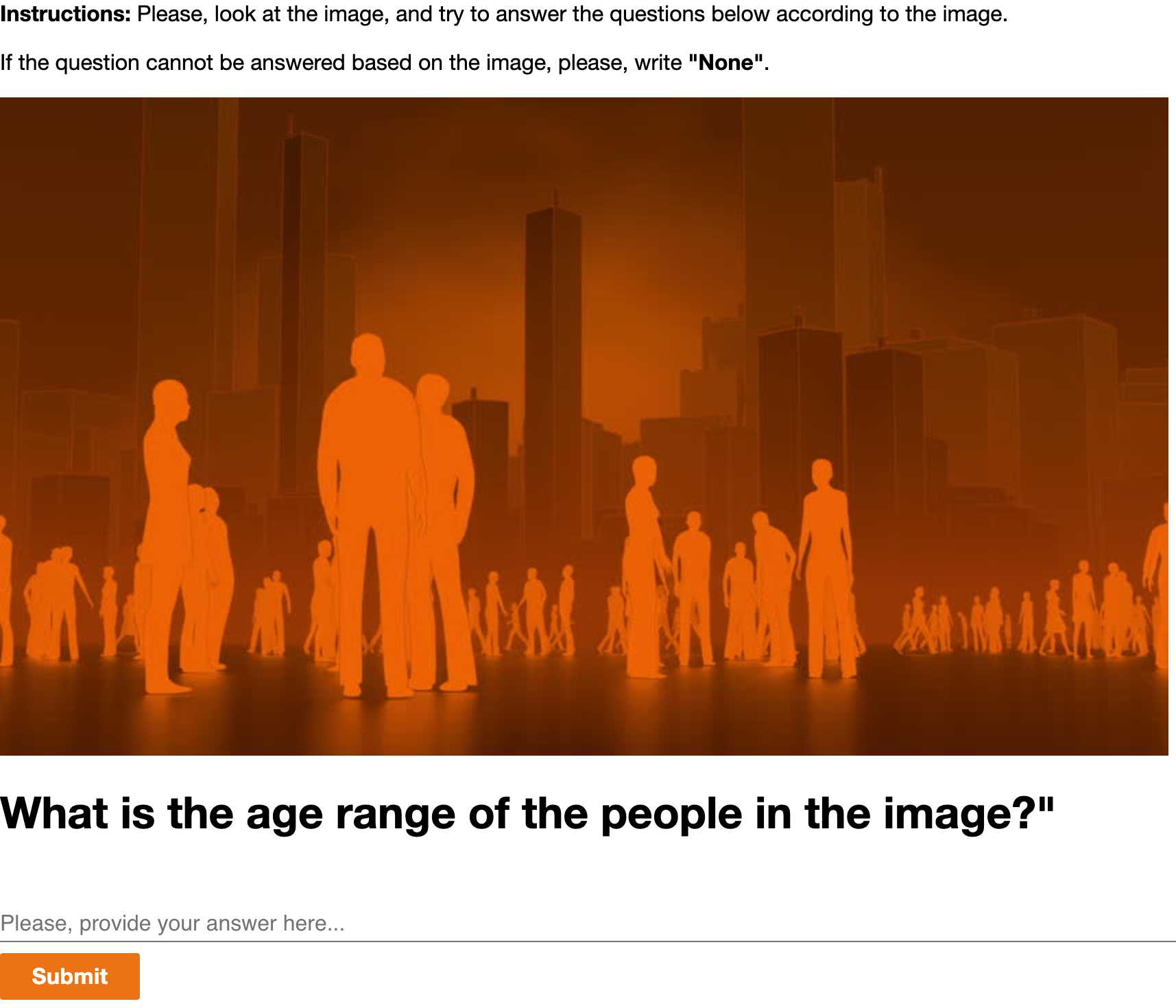}
  \caption{Human evaluation interface of the question-answerable experiments in Tab.\ref{tab:uncertainty}.}
\end{subfigure}
\caption{Human evaluation interfaces}
\end{figure}

\newpage
\subsection{Random Examples of Generated Questions in Conceptual Captioning Dataset}
\noindent 
What is the material of the pier in the image? \\
What is the position of the sign that says "No Swimming Allowed" on the dock? \\
What is the material of the valve? \\
What is the color of the plate on which the cake is placed? \\
What is the expression on the man's face? \\
What is the boy's expression while playing with the sailboat? \\
What is the angle of the camera in the image? \\
What type of flower is it? \\
What is the color of the woman's glasses? \\
What is written on the quote on the mountain? \\
What is the design on the cake? \\
What is the woman's hair color in the image? \\
Are the man and woman standing or sitting in the image? \\
What is the location of the scene depicted in the image? \\
What is the boy's expression? \\
What is the material of the pink running shoes? \\
What is the expression on the man's face? \\
What type of vegetation surrounds the pond in the image? \\
What is the size of the fountain in the image? \\
What is the name of the mountain range seen in the background of the image? \\
What is the name of the park? \\
What is the design of the woman's dress? \\
What is the color of the chainsaw? \\
What is the ethnicity of the two men in the image? \\
What is the woman's pose in the photo? \\
What modifications, if any, have been made to the car in the image? \\
What kind of donuts are in the box? \\
What is the woman's age range in the image? \\
What is the weather like in the image? \\
What is the man's posture like in the image? \\
What kind of lighting is in the room? \\
What is the woman's hair color in the image? \\
What is the woman wearing in the image? \\
What is the woman's pose in the image? \\
What is the type of the lightning bolt? (e.g. cloud-to-ground, cloud-to-cloud) \\
What is the context or setting of the image? \\
What type of event is taking place where the man is performing? \\
What is the shape of the base or stand of the ice sculpture? \\
What is the size of the tent? \\
What is the boy wearing? \\
What is the man doing? \\
What is the material of the sink in the image? \\
What is the age range of the people in the image? \\
What is the style of the hat? \\
What is the background of the image? \\
What is the surrounding environment like in the image? \\
What is the woman wearing? \\
What is the shape of the backyard? \\
What is the material of the backpack the man is carrying? \\
What is the texture of the granny square? \\
What is the style of the bridge? \\
What is the size of the sun in the image?

\newpage
\subsection{More Qualitative Results}

\begin{figure}[b]
\centering
\includegraphics[width=\textwidth]{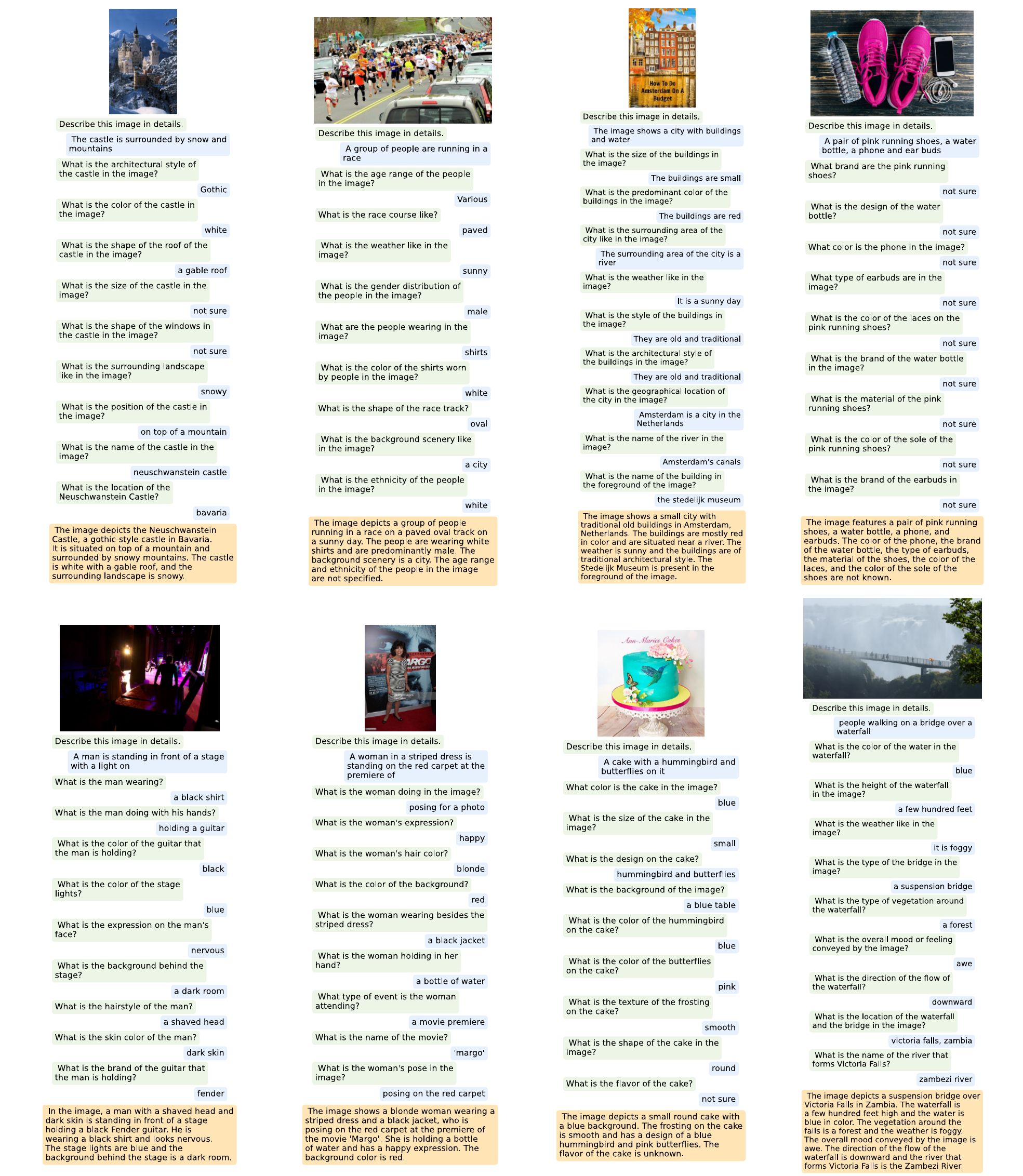}
\caption{Random samples from CC \cite{sharma2018conceptual}.}
\end{figure}

\begin{figure}[b]
\centering
\includegraphics[width=\textwidth]{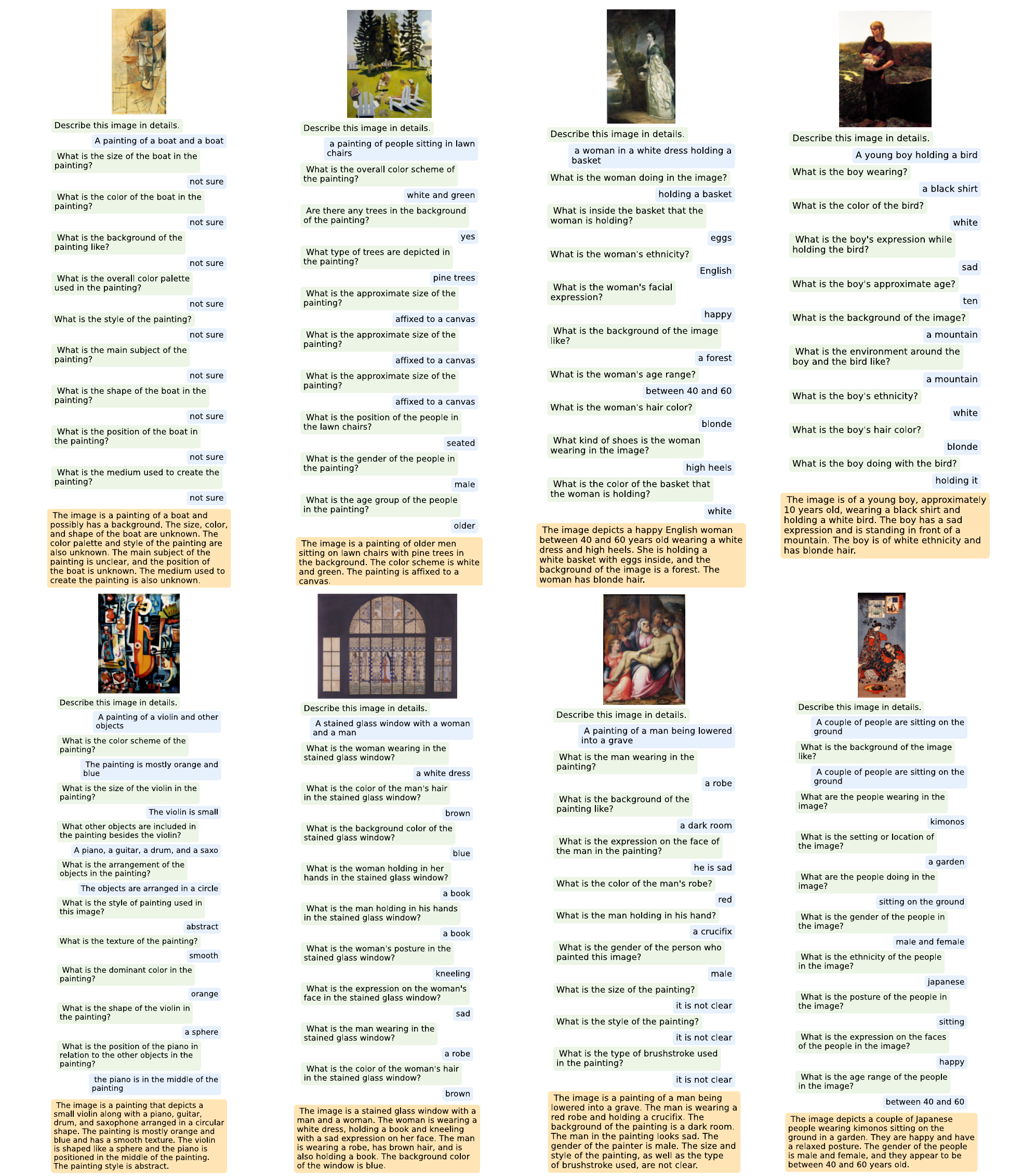}
\caption{Random samples from WikiArt \cite{saleh2015large}.}
\end{figure}

\begin{figure}[b]
\centering
\includegraphics[width=\textwidth]{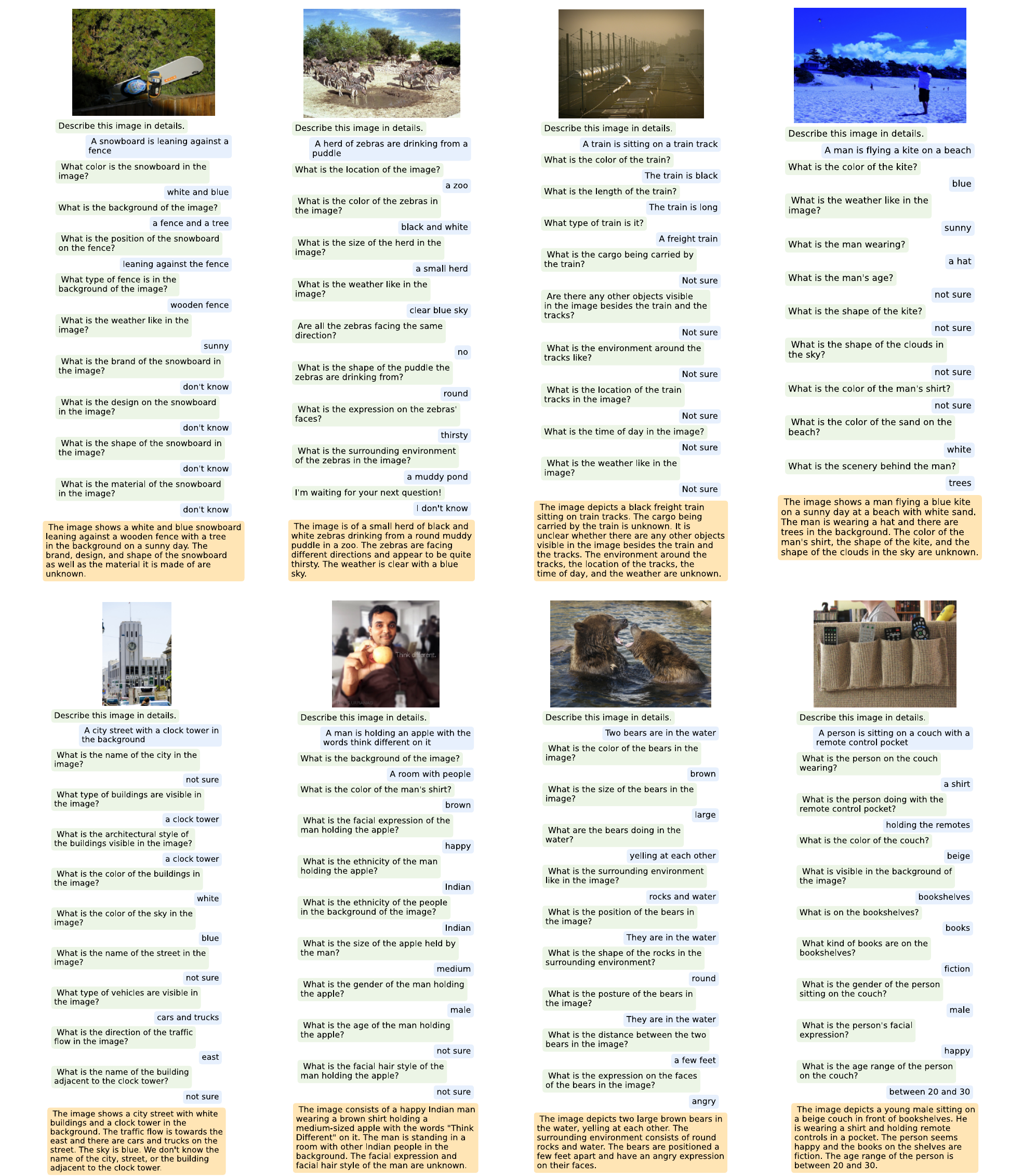}
\caption{Random samples from COCO \cite{lin2014microsoft}.}
\end{figure}

\twocolumn

\end{document}